\documentclass[sn-mathphys-num]{sn-jnl}

\usepackage{graphicx}%
\usepackage{multirow}%
\usepackage{amsmath,amssymb,amsfonts}%
\usepackage{amsthm}%
\usepackage{mathrsfs}%
\usepackage[title]{appendix}%
\usepackage{xcolor}%
\usepackage{textcomp}%
\usepackage{manyfoot}%
\usepackage{booktabs}%
\usepackage{algorithm}%
\usepackage{algpseudocode}%
\usepackage{listings}%
\usepackage{graphicx}
\usepackage{subcaption}

\newcommand{\vecX}{\mathbf{x}}
\newcommand{\vecY}{\mathbf{y}}

\theoremstyle{thmstyleone}%
\theoremstyle{thmstyletwo}%

\theoremstyle{thmstylethree}%

\begin{document}

\title[Article Title]{Concealed Adversarial attacks on neural networks for sequential data}


\author*[1]{\fnm{Petr} \sur{Sokerin}}\email{sokerinpo@mail.ru}

\author[1]{\fnm{Dmitry} \sur{Anikin}}\email{Dmitry.Anikin@skoltech.ru}

\author[1]{\fnm{Sofia} \sur{Krehova}}\email{krehovasofia@gmail.com}

\author[1]{\fnm{Alexey} \sur{Zaytsev}}\email{A.Zaytsev@skoltech.ru}

\affil*[1]{\orgdiv{LARSS laboratory, AI Center}, \orgname{Skoltech}, \orgaddress{\street{Bolshoy Boulevard, 30, bld. 1}, \city{Moscow}, \postcode{121205}, \country{Russia}}}




\abstract{
The emergence of deep learning led to the broad usage of neural networks in the time series domain for various applications, including finance and medicine.
While powerful, these models are prone to adversarial attacks: a benign targeted perturbation of input data leads to significant changes in a classifier's output.
However, formally small attacks in the time series domain become easily detected by the human eye or a simple detector model.

We develop a concealed adversarial attack for different time-series models: it provides more realistic perturbations, being hard to detect by a human or model discriminator. 
To achieve this goal, the proposed adversarial attack maximizes an aggregation of a classifier and a trained discriminator loss. 
To make the attack stronger, we also propose a training procedure for a discriminator that provides broader coverage of possible attacks.
Extensive benchmarking on six UCR time series datasets across four diverse architectures --- including recurrent, convolutional, state-space, and transformer-based models --- demonstrates the superiority of our attack for a concealability-efficiency trade-off. Our findings highlight the growing challenge of designing robust time series models, emphasizing the need for improved defenses against realistic and effective attacks.
}




\maketitle

\section{Introduction}\label{sec:intro}
Time series data appear in a wide range of applications, from healthcare and finance to energy systems and environmental monitoring \cite{marusov2023noncontrastive,zhang2024deep}. 
Time series classification has emerged as a critical task, enabling predictive and diagnostic insights that drive decision-making in these domains \cite{yue2022ts2vec,dempster2020rocket}. 
Neural networks have become a first choice in various such problems, being a powerful solution in this domain \cite{middlehurst2024bake,ismail2019deep} and one of the best non-ensemble approaches.
However, neural networks are known to be more susceptible to adversarial attacks --- subtle data perturbations crafted to deceive models being undetectable to human observers \cite{szegedy2013intriguing}. 
This vulnerability poses significant risks, especially in high-stakes applications like medicine, where incorrect predictions lead to severe consequences.

Adversarial attacks have been studied extensively in image~\cite{akhtar2018threat} and natural language classification~\cite{zhang2020adversarial,goyal2023survey}. Their numerous studies have laid the foundation for understanding adversarial attacks and defenses. 
However, the time series modality presents unique challenges. 
Time series data exhibit complex temporal dependencies, noise characteristics, and domain-specific variations that distinguish it from other modalities~\cite{UCRArchive2018}. 
The key challenge in this domain is ensuring that adversarial perturbations remain concealed, as anomalies in time series data are often easier for humans or machine learning models to detect than in other modalities.

\begin{figure}[thb]
\centering
\begin{subfigure}{.53 \textwidth}
  \centering
  \includegraphics[width=.97\linewidth]{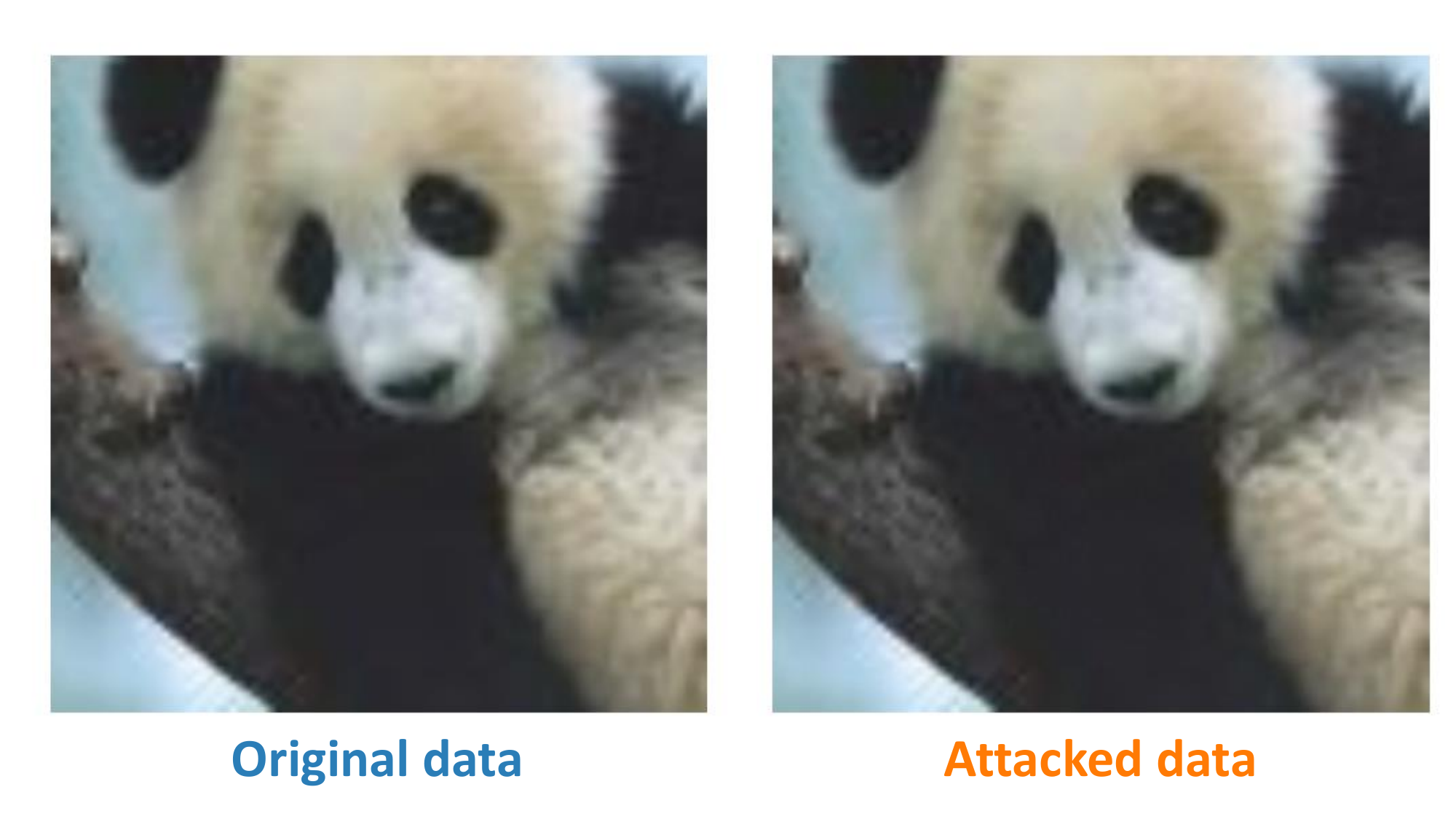}
  \caption{An attack of an image classifier}
  \label{fig:example_cv}
\end{subfigure}%
\begin{subfigure}{.45 \textwidth}
  \centering
  \includegraphics[width=1\linewidth]{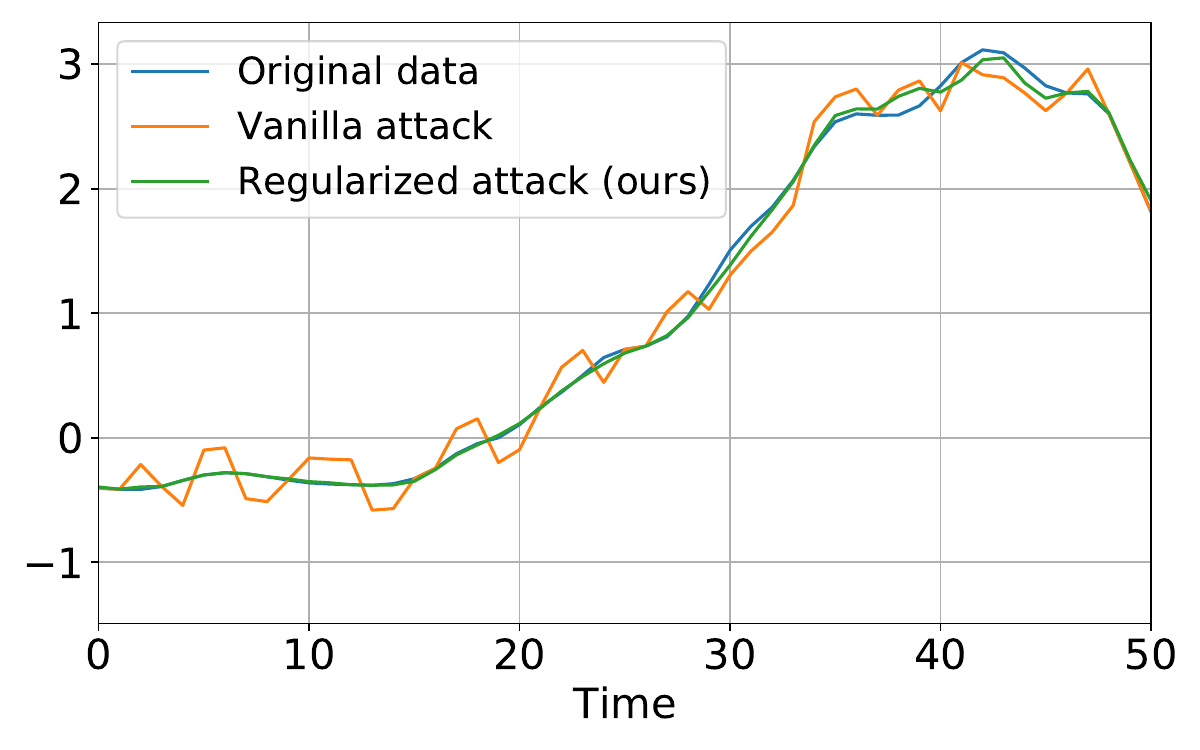}
  \caption{An attack of a time-series classifier}
  \label{fig:example_ts}
\end{subfigure}
\caption{Examples of common adversarial attacks~\cite{fgsm} in the computer vision domain (a) and the time series domain (b). In computer vision, adversarially perturbed and original images appear nearly identical to the human eye. In the time series domain, standard adversarial attacks introduce noticeable artefacts, making them easily detectable. Our regularized approach generates more natural-looking perturbations, enhancing attack concealment.}
\label{fig:example_attack}
\end{figure}

Consequently, we need a more intricate constraint for adversarial perturbation that is wider than common $l_2$ or $l_{\infty}$ norm differences between the initial time series and its attacked counterpart.
The common constraints lead to examples that are easy to detect even by human eyes, as Figure~\ref{fig:example_attack} demonstrates.
Given this definition, we should define a proper attack that resolves the issues with limited concealability. 

This work proposes a concealed adversarial attack properly defined for the time series domain.
In contrast to previous approaches, we can produce realistically looking adversarial examples that are, by design, hard to detect with an additional discriminator.
In more detail, our contributions are the follows: 
\begin{itemize}
 \item A novel concealed adversarial attack. We find out that for the time series domain, the attack should be more realistic to be undetectable to the human eye. So, the proposed method simultaneously looks at the initial model's score and an additional discriminator constructed for attack detection. In time series classification, it is both effective and concealed from human observers and domain-specific anomaly detection. 
 \item An approach to train a discriminator that detects attacks of various strengths. The training starts with strongly perturbed examples; then perturbations gradually decrease to improve the ability to detect small attacks.
 \item A detailed analysis of the inherent characteristics of time series data that make traditional adversarial attacks easily detectable, contrasting this with the challenges observed in image-based attacks. We consider four different models, three attacks and six diverse datasets, e.g. Coffee, FreezerRegularTrain, GunPoint, GunPointMalevsFemale, PowerCons, and Strawberry. The model types include Transformers, Convolutional Neural Networks, Recurrent Neural Networks and State-Space models. The attacks are two white-box attacks: an iterative Fast Gradient Sign Method (iFGSM), Projected Gradient Descent (PGD), and a black-box attack, SimBA.
 We compare the proposed attack to reasonable alternatives, showing that it has the best empirical quality. 
 These findings also present a more detailed view of how adversarial attacks should look in the time series domain.
 \end{itemize}

To facilitate reproducibility and further research, we provide full implementation details, including code and datasets, on our \href{https://github.com/petrsokerin/Hidden_adversarial_attacks/tree/main}{GitHub} page.

\section{Related works}\label{sec:related_work}

Time series classification, a critical task within sequential data processing, has made notable progress in part due to the emergence of deep learning. 
However, adversarial attacks discovered in computer vision (CV) have extended into the deep time series models. This section overviews neural time series classifiers and adversarial attacks and defenses, from their origins in computer vision to their extension into sequential data.
The conclusion highlights the specific challenges for concealed attacks in the time series domain.

\paragraph{Time series classification}

Time series classification (TSC) is a core task within time series analysis~\cite{yang2006challenging}, crucial for various diverse applications, including finance and healthcare. 


Traditional TSC approaches historically dominated the field, relying on feature engineering and classical machine learning. Such methods as One Nearest Neighbor with Dynamic Time Warping \cite{keogh2005exact}, as well as more sophisticated methods, including COTE \cite{bagnall2015time} and Weasel \cite{schafer2017fast}, were the SOTA sequential classification models in the last decade. 
Building on these foundations, more recent advancements, such as HIVE-COTE 2.0 (HC2) \cite{middlehurst2021hive}, have further pushed the boundaries of time series classification. This ensemble approach consists of four main components: Temporal Dictionary Ensemble (TDE), Diverse Representation Canonical Interval Forest (DrCIF), The Arsenal, and Shapelet Transform Classifier (STC). 


While HC2 remains one of the most advanced hybrid methods for time series classification \cite{middlehurst2024bake}, deep learning approaches have gained traction for their ability to extract complex features and scale effectively \cite{mohammadi2024deep}, working well for many real-world applications. However, no single deep learning architecture has emerged as the leading approach, with different models excelling in different scenarios.
Convolutional neural networks (CNNs) quickly became popular, with models like ResCNN \cite{rescnn}, which integrated residual connections from ResNet \cite{he2016deep}, improving gradient flow and multi-scale feature extraction. 
Recurrent architectures, including RNNs and LSTMs \cite{10.1162/neco.1997.9.8.1735}, were also widely adopted, particularly for capturing long-term dependencies, with hybrid approaches like RNNAttention \cite{tsai} combining recurrence with attention mechanisms.
More recently, transformer-based models have become popular, with PatchTST \cite{nie2022time} emerging as a state-of-the-art approach. It segments time series into patches and applies self-attention to capture long-range dependencies efficiently. State space models (SSMs) have also emerged as a promising direction, with S4 \cite{gu2021efficiently} being a key example. S4  reparameterizes structured state matrices, enabling efficient handling of lengthy sequences and flexible switching between recurrent and convolutional representations. 
These models demonstrate strong performance across diverse problems while excelling under specific architectural constraints.




Benchmark repositories like the UCR Time Series Classification Archive \cite{UCRArchive2018} and the UEA Multivariate Time Series Archive \cite{bagnall2018ueamultivariatetimeseries} have greatly facilitated TSC model development. These repositories offer diverse datasets across multiple domains. The UCR archive, for instance, includes over 100 datasets covering tasks like human activity recognition, ECG classification, and sensor data analysis. Based on the current benchmarking landscape, we selected six diverse datasets for univariate time series classification.

\paragraph{Adversarial attacks}
Deep neural networks, despite their ability to handle complex patterns, are susceptible to small data perturbations that result in a wrong prediction \cite{szegedy2013intriguing}, called adversarial attacks. 
Szegedy et al. \cite{szegedy2013intriguing} were among the first to demonstrate that adding subtle, imperceptible noise to images could drastically alter predictions. This discovery prompted extensive research into attack strategies and defense mechanisms \cite{wang2023adversarial,zaytsev2023designing}.

Adversarial attacks generally fall into white-box or black-box categories. White-box attacks assume full access to the model’s parameters and architecture \cite{tramer2017ensemble}. Early work introduced the Fast Gradient Sign Method (FGSM) \cite{fgsm}, which uses a single gradient step to perturb inputs. Its iterative variant (iFGSM) \cite{ifgsm} repeats the process multiple times for an attack. Projected Gradient Descent (PGD) \cite{madry2017towards} further enhances iFGSM by projecting each perturbation  into an 
$l_p$-norm ball, targeting model corruption more successfully.
Black-box attacks have no access to model internals and rely on query-based strategies. A strong Simple Black-box Adversarial Attacks (SimBA) \cite{guo2019simple} iteratively add random vectors from a predefined orthonormal basis, guided solely by model output. This attack has proven highly successful in real-world environments.

We adopt iFGSM, PGD, and SimBA because they are straightforward, competitive, and readily adaptable to time series. While domain-specific methods continue to evolve, these established attacks remain competitive baselines for a broad range of applications.

\paragraph{Adversarial attacks in time-series domain}

Adversarial attacks have received considerable attention in computer vision (CV) for their capacity to manipulate image classifiers. However, despite the growing use of deep learning models in TSC, adversarial attacks remain relatively underexplored in this field~\cite{Ding_Zhang_Feng_Huang_Jiang_Yang_2023}. 

White-box adversarial attacks, which use gradients  \cite{akhtar2018threat} \cite{tramer2017ensemble}, expose model vulnerabilities across domains. In \cite{ifgsm}, Fawaz demonstrates the effectiveness of straightforward iFGSM and the Basic Iterative Method (BIM) on time series. 
However, iterative methods create perturbations that are easily noticeable, highlighting concealability as a key challenge in time-series adversarial attacks.

Beyond gradient-based approaches, researchers have begun developing attack methods explicitly tailored to time series data. Karim et al. \cite{karim2020adversarial} introduced adversarial transformation networks, which train a neural network to generate adversarial examples. Later, Harford et al. \cite{harford2020adversarial} extended this approach to multivariate time series, showing its broader applicability.

While adapting existing CV-based white-box methods like FGSM and BIM to time series data has proven effective, the detectability of perturbations and developing time series-specific attacks remain active research challenges.

\paragraph{Detecting attacks in time-series domain and hidden attacks}
In \cite{abdu2020detecting}, authors present a method targeting FGSM and BIM attacks by treating identifying adversarial samples as an outlier detection problem. The approach builds a normality model using information-theoretic and chaos-theoretic measures, achieving up to $97\%$ detection accuracy on datasets from the UCR Time Series Archive.

This highlights a central challenge in the field: attack concealment. Designing adversarial perturbations that not only mislead models but also remain imperceptible to human observers or detection mechanisms is particularly difficult in time series data, where minor modifications often result in abrupt, easily noticeable changes \cite{pialla2022smooth}.

To address this, \cite{wang2022tsfool} proposes a multi-objective optimization approach incorporating a “Camouflage Coefficient,” which balances attack efficiency with the sample’s proximity to its original class, improving concealability. Similarly, \cite{4fea73ee897f468da2d0e8e0e810d1b8} introduces a fused LASSO-type regularization to smooth adversarial perturbations and combines it with a Kullback-Leibler divergence loss to retain attack efficacy. While the authors highlight that neural network models can detect adversarial perturbations, their approach does not use a discriminator to optimize concealability.

\paragraph{Research gap}
The adversarial attacks on neural time series classifiers remain detectable and require further research on their concealment. 
Introducing an additional discriminator can lead to more realistically-looking attacks without compromising their efficiency. 
Moreover, additional research on discriminators as a defense is required to better understand how protected existing models could be.

\section{Methodology}\label{sec3}
\subsection{Problem statement}
We consider a time series classification model $f(\vecX)$ that takes a multivariate time series $\vecX$ as an input.
Our objective is to design a concealed adversarial attack $h$ that manipulates the predictions of a target model $f(\vecX)$ while being difficult to detect. To achieve the second goal, our method uses an additional discriminator $D$ that aims to detect adversarial examples.
More formally:
\begin{itemize}
    \item \textbf{Target Model $ f $}: The model that the adversarial attack aims to deceive, to generate perturbations that mislead its predictions.
    
    \item \textbf{Discriminator Model $D$}: A model trained to distinguish adversarial data from original data. The aim is to create perturbations that are hard for $D$ to detect.
    
    \item \textbf{Concealed Adversarial Attack $h$}: A function that generates perturbations on the input $\vecX$ to deceive the target model $f$ while ensuring that the perturbations remain subtle and undetectable by the discriminator $D$. To account for both components, the attack uses an aggregation function $g$ for outputs of $f$ and $D$.
\end{itemize}

\subsection{Concealed attack construction}

\begin{figure}[thb]
\centering
\begin{subfigure}{.5\textwidth}
  \centering
  \includegraphics[width=1.1\linewidth]{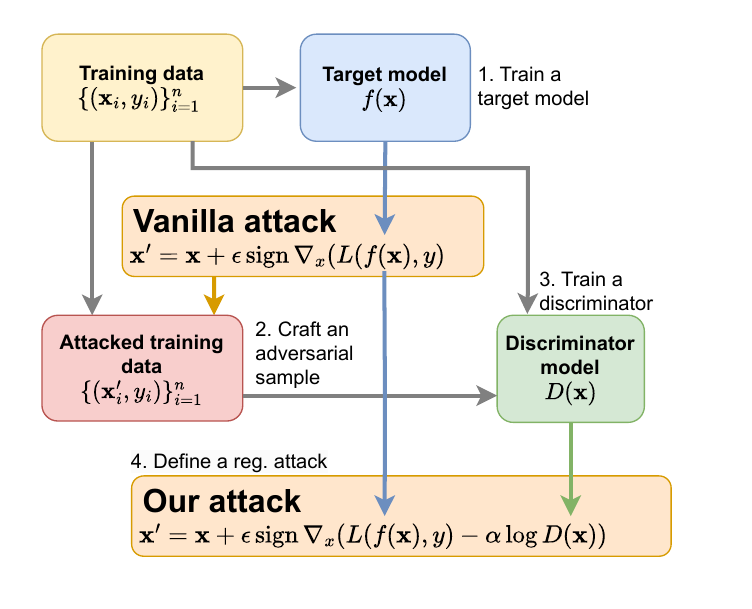}
  \caption{Attack design scheme}
  \label{fig:attach_scheme}
\end{subfigure}%
\begin{subfigure}{.5\textwidth}
  \centering
  \includegraphics[width=1.0\linewidth]{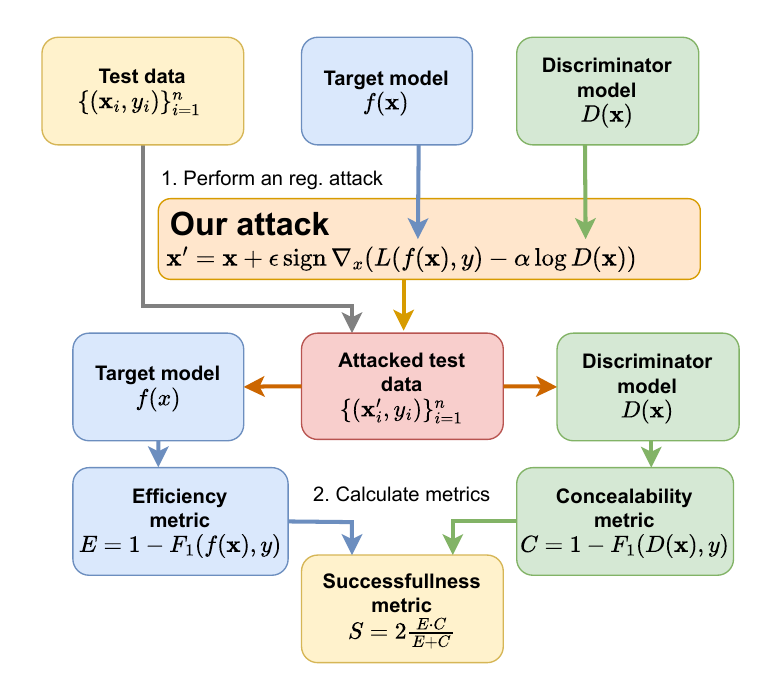}
  \caption{Attack validation scheme}
  \label{fig:validation_scheme}
\end{subfigure}
\caption{Pipeline of definition (a) and validation (b) for an adversarial attack with the discriminator regularization. 
Firstly, we train the target model for an attack. Then, we generate adversarial data with a vanilla attack to train discriminator models to classify whether the data object has been attacked or not.
The next step is to apply a final attack on the original data to generate perturbed data objects to compromise both the target model and the discriminator. Finally, we get Efficiency and Concealability metrics to estimate the results of our final attack.}
\label{fig:pipeline}
\end{figure}

We design a pipeline for constructing a concealed adversarial attack depicted in Figure~\ref{fig:attach_scheme}.
It consists of the following steps:
\begin{enumerate}
    \item Generate an adversarial dataset by applying an adversarial attack with low magnitude. Objects from this adversarial dataset, labelled with 1, and the original data, labelled with 0, are merged in a single dataset $B_D$ for the discriminator model $D$ training. 
    \item Train a discriminator model $D$ on adversarial samples and samples from original dataset $B$.
    \item Define a discriminator-based adversarial attack.
\end{enumerate}

The subsequent subsections discuss in detail all steps of the pipeline. 
We begin by describing a target classifier and a vanilla adversarial attack. 
Then, we present our methodological contributions: the training scheme for a discriminator model and our discriminator-regularized concealed adversarial attack.


\paragraph{Classification models for an attack}


The classification target model $f$ is trained on dataset $B = \{(\vecX_i, y_i)\}_{i = 1}^n$, including features $\vecX_i$ and target labels $y_i$ for an object $i$. 
The entire dataset includes feature matrix $X = \{ \vecX_i\}_{i = 1}^n$ and label vector $\vecY = \{ y_i\}_{i = 1}^n$.
The dataset is split into the train part ($X_{\mathrm{train}}$, $\vecY_{\mathrm{train}}$) and the test part ($X_{\mathrm{test}}$, $\vecY_{\mathrm{test}}$). 
The model predictions $f(X_{\mathrm{test}})$ are compared with ground true labels $\vecY_{\mathrm{test}}$ to estimate quality metrics. 
We train models using a common cross-entropy loss $L( f(X_{\mathrm{train}}), \vecY_{\mathrm{train}})$. 

In the field of time-series classification, there is no universally dominant model. 
While Transformer models, which excel in NLP, can be applied to time-series classification tasks, their effective training can be challenging due to the limited availability of large pre-training datasets. 
Convolutional Neural Networks (CNNs), Recurrent Neural Networks (RNNs) and State Space Models also remain popular choices. In our study, we employ the convolutional ResCNN model \cite{rescnn}, the recurrent RNNAttention model \cite{tsai}, the Transformer-based model PatchTST \cite{nie2022time}, and a state-space models S4 \cite{gu2021efficiently}.

\subsection{Adversarial attacks}

Adversarial attacks are methods for compromising target model $f$ predictions.
Mainly, adversarial attacks $h$ add special adversarial perturbations $h(\vecX)$ to original data object features $\vecX$ to get perturbed data object $\vecX^{\prime} = \vecX + h(\vecX)$.
If an attack runs over several iterations,
we start with $\vecX^0 = \vecX$ and iteratively update it by generating $\vecX^{t + 1} = \vecX^{t} + a(\vecX^t)$.
We obtain our adversarial example at the last $T$-th iteration: $\vecX^{\prime} = \vecX^T$.
Below, we consider different options for defining $T$ and $\vecX^T$.

Our paper covers several frequently used attacks. 
The main requirements for an attack are Efficiency in other domains and simplicity in potential regularization. 
We select two gradient-based attacks, iFGSM and PGD, and a black box attack, SimBA. The detailed definitions for them are in Appendix~\ref{sec: attacks}.
As an example of the attack, we use an iterative attack, iFSGM. It has the following form at $t$-th iteration:
\begin{align*}
    \vecX^{t + 1} = \vecX^{t} + \epsilon \operatorname{sign} \left( \nabla_{\vecX^t} L \left(f(\vecX^t), y\right) \right),
\end{align*}
where $\epsilon$ is the attack step size.

\subsection{Discriminator model}

\paragraph{Discriminator for adversarial attacks}

Our discriminator $D$ is trained to distinguish between normal $\vecX$ and adversarially perturbed objects $\vecX' = \vecX + h(\vecX)$.

To train a discriminator, we need training data $B_D$. 
It is composed of two subsamples: a subsample of initial training sample $B^0 = \{(\vecX_i, 0)\}_{i = 1}^n$ with normal $\vecX$ and a sample $B^1_{\theta} = \{(\vecX'_i, 1)\}_{i = 1}^n$ of adversarially perturbed objects $\vecX' = \vecX_i + h_{\eta}(\vecX_i)$.
The parameters $\theta$ define a specific attack and constitute, in most cases, the number of iterations $T$ and the attack step size $\varepsilon$. 
The labels $0$ and $1$ mark if an attack perturbed an object.

The discriminator model is used for two aims: 
\begin{itemize}
    \item \emph{Metric of Concealability.} The errors of the discriminator during an attack signify the quality of an attack.
    \item \emph{The tool to make the attack more hidden.} The discriminator can serve as a regularizer to increase the concealability of an attack. 
\end{itemize}




\paragraph{Robust discriminator training}

Vanilla training of a discriminator typically results in a weak classifier because we don't learn it to recognize the diverse attacked options with varying attack power and the number of steps. 
To overcome this limitation, we provide a more diverse and comprehensive training set of adversarial examples. 
Additionally, following the curriculum learning, the attack amplitude gradually decreases during training, increasing the problem's complexity.

The steps for the discriminator training are in Algorithm~\ref{alg:adv_concealment}.
The thresholds for the minimal accuracy of the discriminator $D(\vecX)$ and reduction of the attack strength in the algorithm are $0.9$ and $0.8$ correspondingly. They suit all considered samples and models. For some models and datasets, we need only $3-4$ iterations to achieve chosen criteria, for others, we train the discriminator with $7-8$ iterations.

\begin{algorithm}[h!]
\caption{Iterative Robust Discriminator Training}
\label{alg:adv_concealment}
\begin{algorithmic}[1]
    \Require Clean data $B^0$, initial attack strength $\varepsilon_{\mathrm{init}}$, total attack iterations $T$
    \State initialize $\varepsilon = \varepsilon_{\mathrm{init}}$ \Comment{ \emph{$\varepsilon_{\mathrm{init}}$ is typically large}}
    \State \textbf{generate adversarial sample:} Apply an attack with the strength $\varepsilon$ and $T$ iterations to $B^0$ to obtain $B^1_{\mathrm{gen}} = B^1_{\varepsilon}$.
    \State \textbf{train a discriminator $D(\vecX)$:} Use $B^0$ and $B^1_{\mathrm{gen}}$ \Comment{\emph{The initial discriminator accuracy is typically $\sim 1$.}}
    \While{accuracy of $D(\vecX) > 0.9$}
        \State reduce the attack strength $\varepsilon = 0.8 \cdot \varepsilon$
        \State generate a new adversarial sample $B^1_{\mathrm{gen}} = B^1_\varepsilon$
        \State finetune the discriminator using $B^0$ and $B^1_{\mathrm{gen}}$
    \EndWhile \\
    \Return Robust discriminator model $D(\vecX)$
\end{algorithmic}
\end{algorithm}

\subsection{Concealed adversarial attacks}

To make the attack more concealed, we modify a single perturbation step.
Here is an example of a modification for an iFGSM attack:
$$
\vecX^{t+1} = \vecX^{t} + \epsilon  \operatorname{sign}\left[\nabla_{\vecX^t} g \left(L_{\mathrm{target}}\left(f(\vecX^t), y \right), L_{\mathrm{disc}}\left(D(\vecX^t), y_{\mathrm{disc}}\right) \right) \right]. 
$$
Using an aggregation function $g(\cdot, \cdot)$ we take into account both $L_{\mathrm{target}}$ and $L_{\mathrm{disc}}$, that are loss functions of the target model and discriminator respectively. Inside them $y$ and $y_{\mathrm{disc}}$ are true labels for the target task and the discriminator.


Similarly to GANs, the second loss is a log of the discriminator score that we want to minimize. But, unlikely in GAN, we use freezed discriminator model.
So, our step takes the form: 
$$
\vecX^{t + 1} = \vecX^{t} + \epsilon \operatorname{sign}\left[ \nabla_{\vecX^t}g \left(L_{\mathrm{target}}\left( f(\vecX^t), y_{\mathrm{target}}\right), - \log D (\vecX^t) \right) \right].
$$

We consider a simple sum aggregation function:
$$
    g(\vecX) = a(\vecX) + \alpha d(\vecX) = L_{\mathrm{target}} \left( f(\vecX), y \right) - \alpha\log D (\vecX).
$$
It has a weight for the discriminator $\alpha$ and, according to it, takes into account both scores.
Alternatively, our approach can use other aggregation functions, like a harmonic aggregation function that has no hyperparameters to tune: 
\[
    g(\vecX) = 2 \frac{L_{\mathrm{target}} \left( f(x), y \right) \cdot (- \log D (\vecX^t))}{L_{\mathrm{target}} \left( f(\vecX), y \right) - \log D (\vecX) + \gamma}, 
\]
where $\gamma$ is a small value to avoid the zero division problems. 

In our work, we propose other variants of aggregation functions for target and discriminator losses. Namely, we would also consider a hypercone aggregation.
Details on it, as well as a more detailed examination of the properties of possible aggregation functions, are available in Appendix~\ref{sec:app_aggregation}.

\section{Results}\label{sec4}
In this section, we describe the results of experiments on applying adversarial attacks for different models and datasets, starting with a proper validation scheme for the considered approaches.

\subsection{Attack Efficiency and Concealability validation}

During validation, we measure Efficiency of an attack, how concelable it is, and overall aggregated quality, a harmonic mean for them.

Efficiency of the attack we defend is the ability to compromise the target model. So we can get the attack efficiency as $1$ minus $F_1$ score for attack examples $\vecX'$:
$$
E = 1 - F_1 (y_{\mathrm{true}}, f(\vecX')).
$$
Concealability of the attack is the discriminator model error:
$$
C = 1 - F_1 (y_{\mathrm{disc}}^{\mathrm{true}}, D(\vecX)).
$$
Finally, Successfulness $S$ of the attack as the harmonic mean of Efficiency and Concealability:
$$
S = 2 \frac{C \cdot E}{C + E}.
$$
The harmonic mean sharply decreases with an increase of each term $C$ and $E$. 
All three proposed metrics are the greater, the better. 

Given the definition of the quality metrics, we obtain the validation scheme in Figure~\ref{fig:validation_scheme}:
\begin{enumerate}
    \item Perform an attack for the test part of the dataset to generate perturbed data.
    \item Estimate Successfulness, Efficiency, and Concealability metrics for perturbed data. For the estimation, we use the target and discriminator models.
\end{enumerate}

\subsection{Experiments setup and data}

\paragraph{Datasets}
To compare vanilla and concealed adversarial attacks, we attacked models trained on six datasets from the UCR package \cite{UCRArchive2018}: PowerCons, Strawberry, GunPoint, GunPointMalevsFemale, Coffee, and FreezerRegularTrain. These datasets are diverse in data domains and number of data samples. All datasets contain univariate time-series data labelled for the classification task. 
Some datasets are imbalanced, so this is the reason why we take $F_1$ as a basic metric for Efficiency and Concealability calculation. The data for all datasets was divided into training and testing sets as per the original splits. All datasets do not have any time feature and were split by data objects. 

\paragraph{Target model and discriminator training}

All models, ResCNN, RNNAttantion, S4, and PatchTST, were trained on all six datasets. 
We use different model hyperparameters for training models, and $F_1$ measures greater than $0.75$ for all model-dataset combinations. For almost all models, we achieve target metrics higher than $0.9$. 

For discriminator training, we use the same type as the target model: if the target model is ResCNN, the discriminator model is also ResCNN. We trained discriminator models on adversarial training data that was weakly perturbed compared to the Successfullness. Mostly, we decrease the number of attack iterations but not the step of attack size or radius for the PGD attack. We varied different model hyperparameters to achieve better results for discriminator training. In particular, we tuned dropout and learning rates to avoid overfitting. Besides, for the PGD attack, which was hardly detectable for the discriminator, we used smaller models with fewer hyperparameters than in the target model. As was mentioned in \ref{sec3}, we use a special procedure of training the discriminator, decreasing the strength of attack five to eight times at $20$\% every time to make the discriminator model more sensitive to small data perturbations. All discriminator models were trained with an accuracy score greater than $0.85$ on small perturbations. The accuracy metric is correct in this case because we have an equal number of perturbed and original data samples. For all attacks, iFGSM, SimBA, PGD, and SGM, we train separate discriminator models using vanilla or non-regularized attack versions.

\paragraph{Experiments setup}{}
All discriminators were trained on the vanilla attack without regularization. We set the number of steps to $10$ for the iFGSM and PGD attack and $500$ steps for the SimBA Attack. We begin training the discriminator with the next parameters of the attacks: the $\epsilon$ was set to $0.03$ for the iFGSM attack, $e$ was set to $300$ for SimBA, and $\eta$ was set to $0.3 $for the PDG attack. We decrease these parameters 5-8 times on 20\% every time to provide iterative discriminator training. 

For various types of attacks, hyperparameters were optimized through grid search. Specifically, the iFGSM attack with regularization and sum aggregation was applied using different $\epsilon$ values: $0.005, 0.01, 0.03$, and $0.05$, along with regularization $\alpha$ values of $0.001, 0.01, 0.1, 1, 10$, and $100$. For hypercone aggregation, the $\Delta$ parameter was varied across the values $-0.5, -0.3, -0.1, 0, 0.1, 0.3, 0.5$, and $1.0$. The Projected Gradient Descent (PGD) attack was applied with different $\eta$: $0.05, 0.1, 0.25, 0.5$, and $1$. The PGD attack with regularization employed the same $\alpha$ values as the iFGSM attack with regularization.

We also add an SGM attack using KL divergence and $L_2$ distance from original data for regularization \cite{4fea73ee897f468da2d0e8e0e810d1b8}. This method is SOTA concealed adversarial attack for time series data. To train the discriminator, a modified version of the attack was used, with the regularization term coefficients for attack magnitude (L2 distance) and smoothness set to zero while maintaining the noise clipping coefficient, representing the maximum attack magnitude, \(\epsilon = 0.1\), as in the original paper. The attack was then performed with various \(\epsilon\) values: $0.05, 0.1, 0.25, 0.5$, and $1.0$. Additionally, the smoothness and $L_2$ coefficients were assigned equal values, including $0.0$, $0.1$, $0.5$, and~$1.0$.

In estimating attack results, we use the following procedure of results validation: we take the best iteration according to the Successfulness among all attack hyperparameter combinations. For fair metrics estimation, we calculate take metrics after at least 40 attack iterations for the iFGSM and PGD attacks, 1300 iterations for the SimBA attacks, and 400 iterations for the SGM attack or with an Efficiency of more than 0.9 for all attacks. Otherwise, an attack at the first iteration can be taken because of the discriminator's inability to detect small perturbations.

\subsection{Main results}


\paragraph{Ablation Study}

\begin{figure}[thb]
\centering
\begin{subfigure}{.3535\textwidth}
  \centering
  \includegraphics[width=.97\linewidth]{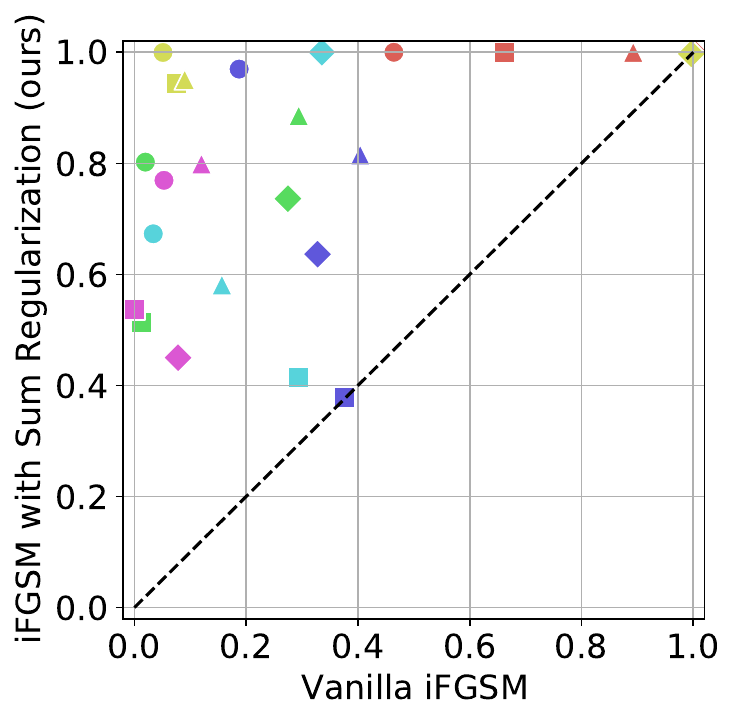}
  \captionsetup{justification=raggedright,singlelinecheck=false}
  \caption{Vanilla vs regularized iFGSM}
  \label{fig:scatter_fgsm}
\end{subfigure}
\begin{subfigure}{.638\textwidth}
  \centering
  \includegraphics[width=.97\linewidth]{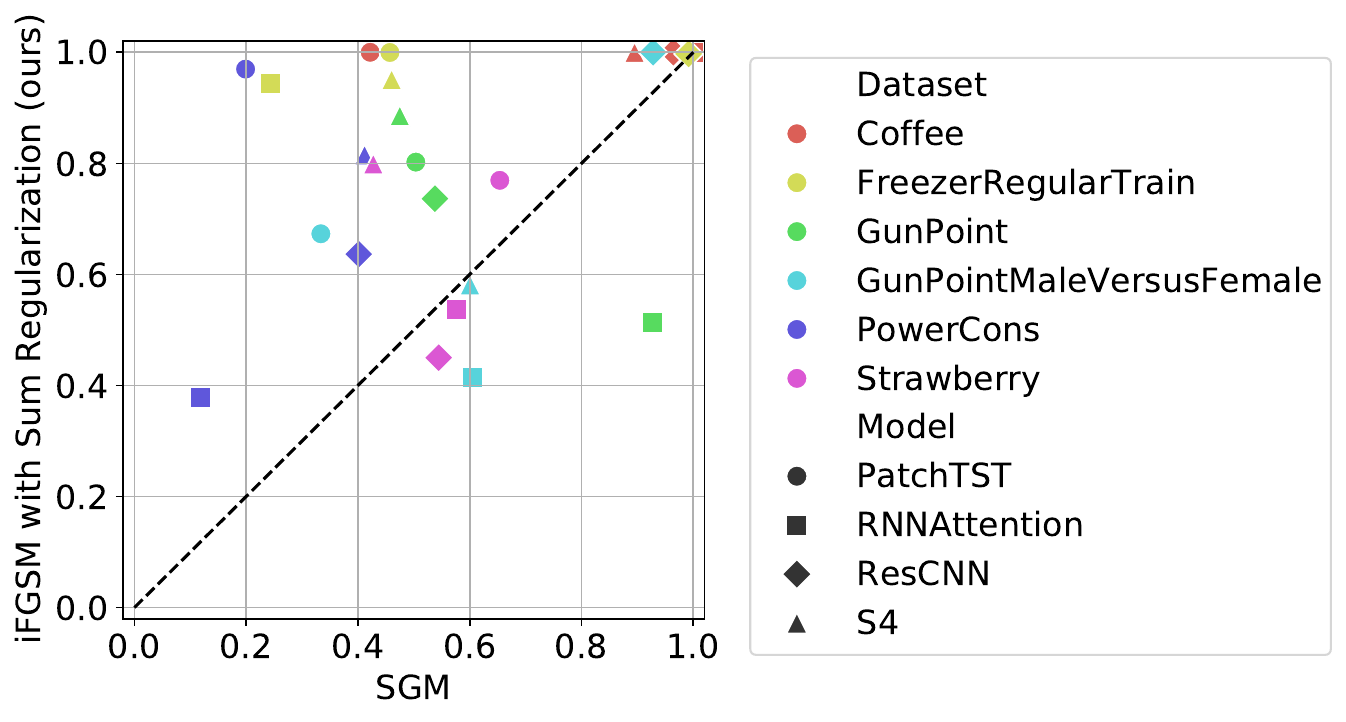}
  \captionsetup{justification=raggedright,singlelinecheck=false}
  \caption{SGM vs regularized iFGSM}
  \label{fig:scatter_kll2}
\end{subfigure}
\caption{The successfulness of iFGSM with sum regularization versus two baseline approaches for different pairs of models and datasets: for almost all experiments, our approach outperforms vanilla attack  (a) and SGM baseline (b).}
\label{fig:scatter_many}
\end{figure}


The aggregated results for all our datasets and models with iFGSM attack are presented in table \ref{tab:mean_rank}. We computed the mean metrics and mean ranks from them for the best iteration of each attack, considering various hyperparameters across all described datasets and models. The best iteration was selected based on Successfulness metrics, while Efficiency and Concealability for this iteration were also reported, following the selection rule outlined in the previous subsection. Additionally, we included the SGM attack for comparison.

\begin{table}[!thb]\centering
\caption{Different iFGSM attack regularization mean rank for Efficiency, Concealability, and Successfullness. Experiments were run on all presented datasets, and ResCNN, RNNAttention, S4, PatchTST models}\label{tab:mean_rank}

    \begin{tabular}{l|ccc|ccc}
    \toprule
    \multirow{2}{*}{Regularization method} &  \multicolumn{3}{c|}{Mean metric ($\uparrow$) } &  
    \multicolumn{3}{c}{Mean rank ($\downarrow$)} \\
     &  Success. &  Efficiency &  Conceal. &  Success. &  Efficiency &  Conceal. \\
    \midrule
    Vanilla attack & 0.300 & \textbf{0.861} & 0.239 & 3.750 & 2.042 & 3.792 \\
    SGM & 0.570 & 0.618 & 0.641 & 2.750 & 3.562 & 2.479 \\
    Harmonic reg. (ours) & \textbf{0.789} & 0.819 & 0.800 & 1.875 & \textbf{1.958} & 2.021 \\
    Sum reg. (ours) & 0.786 & 0.744 & \textbf{0.878} & \textbf{1.625} & 2.438 & \textbf{1.708} \\ 
    \bottomrule
    \end{tabular}

\end{table}

As we can see, a sum regularization performs the best according to the Concealability mean metric, Successfulness and Concealability mean ranks. The Concealability of the sum regularization grows up to $0.878$ compared with $0.239$ for unregularised attacks, while Efficiency drops only to $0.744$ from $0.861$. Harmonic regularization also works well and has better Efficiency than sum regularization but worse Concealability. The Harmonic regularization method is the best according to the mean Successfullness but the second best according to the mean rank. Both harmonic and sum regularization outperform an SGM adversarial attack for all three metrics. More detailed results are presented in Appendix \ref{secA1}.

We also compare our iFGSM regularised attack side by side with vanilla iFGSM attack and SGM baseline. We visualize Successfulness of iFGSM regularization on the y-axis and vanilla iFGSM (a) and SGM attacks (b) on the x-axis in Figure~\ref{fig:scatter_many}. As we can see, all points are located higher than the diagonal line. Our sum regularised attack often performs better than vanilla iFGSM in most experiments. For a small number of cases, regularized iFGSM is worse than SGM, but for most experiments, our method outperforms SGM.  


\paragraph{Sensitivity Study}

\begin{figure}[thb]
    \centering
    \includegraphics[width=0.7\textwidth]{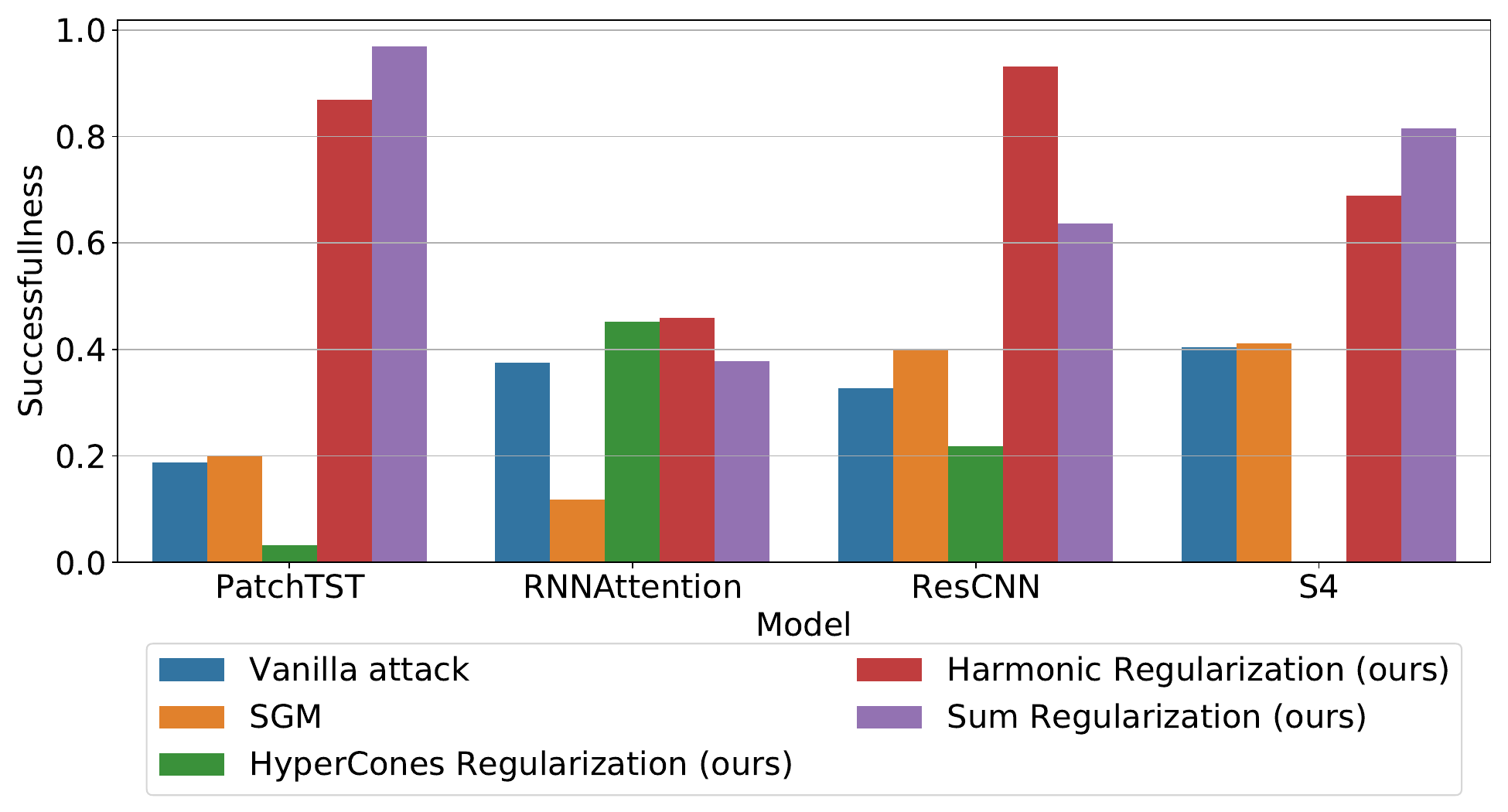}    
    \caption{Successfullness for different models (the greater, the better). The performance of different regularization types varies depending on the model's architecture.}
    \label{fig:model_means}
\end{figure}

Our next step was to compare attack regularization across various datasets. We conducted experiments using all three types of regularization for the iFGSM attack across four models on the PowerCons dataset. As shown in figure \ref{fig:model_means}, both the transformer model PatchTST and the state-space model S4 were successfully attacked using harmonic and sum regularization but remained resilient to hypercones regularization. However, in recurrent architecture RNNAttention, hypercones regularization outperformed both harmonic and sum regularization. Additionally, harmonic regularization proved to be more effective than sum regularization for the convolutional model RessCNN. Notably, hypercones regularization performed poorly across all models except for the RNNAttention model.
Furthermore, in all our experiments, both harmonic and sum regularization consistently outperformed the SGM attack. Based on these findings, we will focus solely on harmonic and sum regularization in subsequent experiments. These results also highlight that different regularization techniques are more effective for specific model architectures.

\begin{figure}[ht]
\centering
\begin{subfigure}{.44\textwidth}
  \centering
  \includegraphics[width=.97\linewidth]{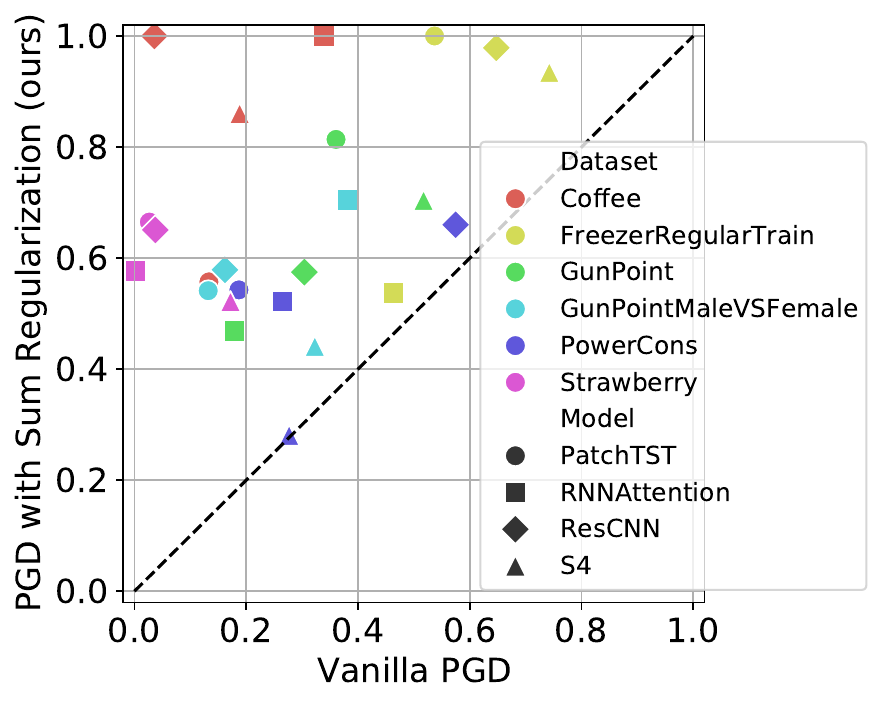}
  \caption{PGD attack}
  \label{fig:scatter_fgsm}
\end{subfigure}
\begin{subfigure}{.55\textwidth}
  \centering
  \includegraphics[width=.97\linewidth]{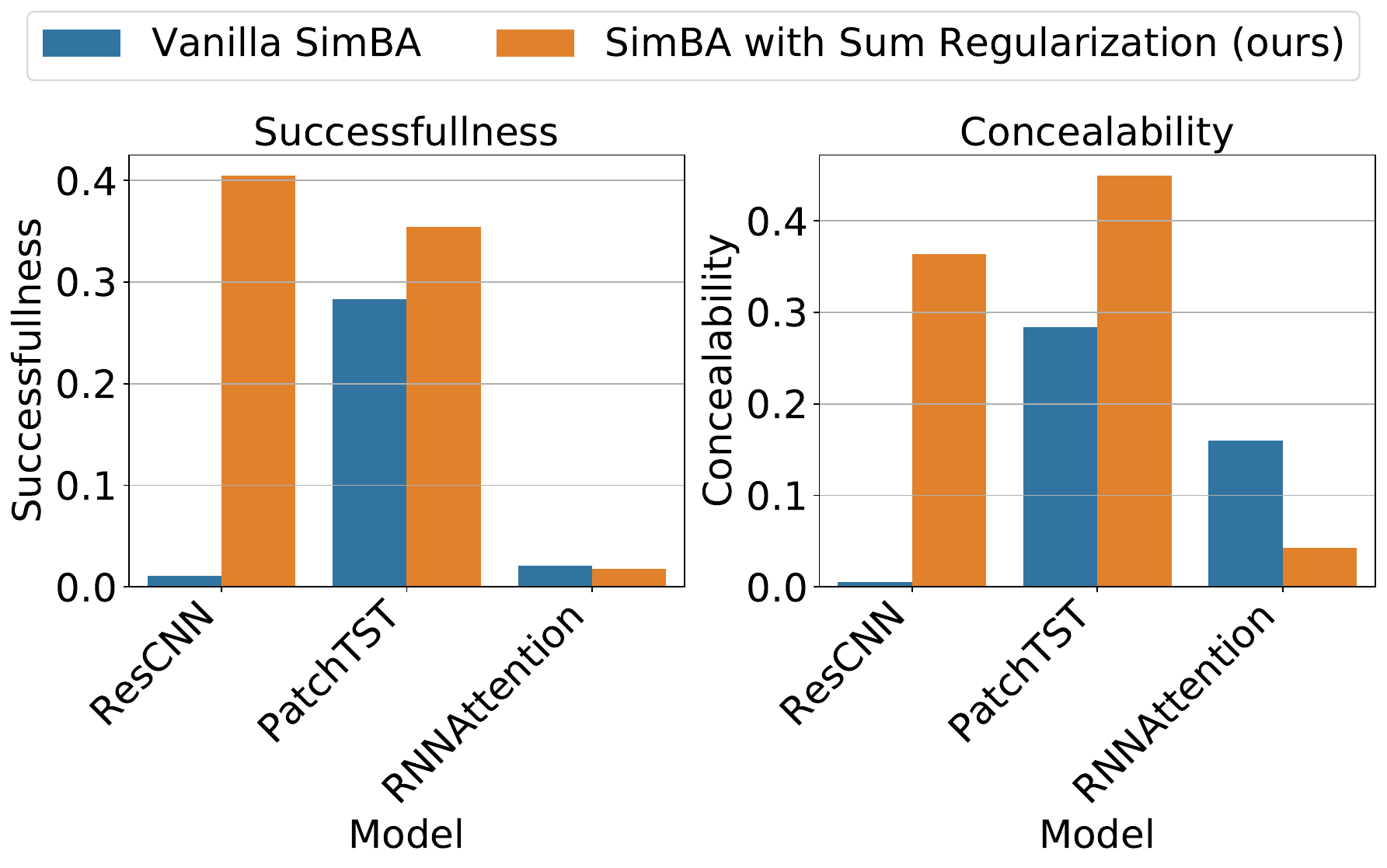}
  \caption{SimBA attack}
  \label{fig:simba_res}
\end{subfigure}
\caption{The iFGSM with sum regularization for almost all experiments outperforms vanilla attack version (a), and SGM baseline (b).}
\label{fig:pgd_simba}
\end{figure}

Our method performs well against other gradient-based attacks. As provided in Figure \ref{fig:scatter_fgsm}, the PGD attack with Sum regularization outperforms vanilla PGD, similar to our findings with FGM. However, iFGSM with Sum or Harmonic regularization yields even better results compared to PGD, as shown in Table \ref{tab:all_metrics}.  

On the other hand, our adversarial attack regularization method performs poorly with the SimBA attack, as shown in figure \ref{fig:simba_res}. The non-gradient nature of SimBA makes training the discriminator for effective attack detection and regularization significantly more difficult. It appears that because SimBA perturbs the input in a highly random manner—unlike gradient-based methods—it prevents the discriminator from identifying distinguishable patterns. Additionally, for some models, SimBA requires an excessive number of iterations to achieve an effective attack, which further undermines the results of the experiments.


Last but not least, we visualize data attacked with vanilla iFGSM adversarial and regularized attacks. 
The plots are in Figure \ref{fig:example_ts} from the Appendix~\label{sec:data_samples}, as well as technical details on experiments. 
Visually, the regularized attack results in fewer artefacts, with the object closely resembling the original.


\section{Conclusion}\label{sec5}
In this study, we address the challenge of concealed adversarial attacks on sequential data models. Our novel approach utilizes a discriminator model trained on a binary classification task to determine whether input data has been adversarially perturbed. This discriminator serves a dual function: it operates as both an adversarial perturbation detector and a regularization mechanism for the attack. The principal contribution of this work is the development of advanced adversarial attack regularization techniques that enhance the concealment of attacks while preserving their Efficiency. By incorporating discriminator loss into conventional adversarial mechanisms, our approach achieves a crucial balance between attack efficiency and detectability, thereby addressing a significant gap in the existing research. Furthermore, we illustrate the inherent visibility of adversarial attacks in time series models and introduce an effective methodology for training discriminator models. Our analysis reveals that standard adversarial attacks, such as iFGSM and PGD, introduce noticeable perturbation artefacts that can be easily detected. In contrast, our regularization techniques significantly suppress these artefacts, making attacks harder to distinguish from original data.

The proposed experimental framework integrates the training of target models, the generation of adversarial datasets, and the use of discriminator-based regularization. It provides a comprehensive approach to assess both the Efficiency and Concealability of adversarial attacks. As part of this framework, we explore various strategies for discriminator regularization and conduct an in-depth comparison of their performance across different models and datasets. Specifically, we assess three regularization techniques: Sum Regularization, Harmonic Regularization, and Hypercones Regularization — using four deep learning architectures: ResCNN, RNNAttention, S4, and PatchTST.

Our experimental results demonstrate that Sum and Harmonic Regularization consistently yield the most favourable outcomes in terms of both attack Efficiency and Concealability. Notably, both techniques outperform the SGM adversarial attack, which is also designed to be less detectable across all three key evaluation metrics — Success, Efficiency, and Concealability - further validating their Efficiency. However, while our approach performs strongly against multiple gradient-based attacks, it does not generalize well to blackbox methods, such as SimBA.

Looking ahead, the insights derived from this study can inform the development of more resilient time series classification models and foster further exploration into advanced adversarial defense mechanisms. Future research should also focus on refining adaptive regularization strategies that dynamically adjust based on the characteristics of the target model and dataset. Also, extending our approach to non-gradient-based attacks remains a critical challenge. Finally, investigating methods that integrate gradients from multiple models — whether sharing the same architecture or not — could significantly enhance concealability while maintaining attack Efficiency due to improved generalization capabilities. 

\bibliography{sn-bibliography.bib}

\newpage
\appendix
\begin{appendices}
\section{Adversarial attacks algorithms}\label{sec: attacks}

This section contains a detailed description of adversarial attack algorithms used in our research. 
These attacks are described below.
\begin{itemize}
    \item \textbf{iFGSM attack}. Several papers demonstrate the Efficiency of this attack in the CV domain. Besides, the iFGSM attack was also applied to sequential data successfully. The basic attack approach for this method is the fast gradient sign method (FGSM). The idea is to make perturbations with the sign of the model loss $L$ gradient. The sign of gradient for every data object $\vecX$ is multiplied by the strength of an attack hyperparameter $\epsilon$ to get perturbed data object $\vecX^{\prime}$.
    \begin{align*}
        \vecX^{\prime}=\vecX+\epsilon\operatorname{sign}\left(\nabla_{\vecX} L(f(\vecX), y)\right),
    \end{align*}
    where  $\nabla_{\vecX} L(f(\vecX), y)$ --- the gradient of the model.
    

    
    However, the main disadvantage of a vanilla FGSM attack is that it transforms data only once. It cannot be enough to change the predictions of the model. So, the iterative modification of this approach was proposed \cite{ifgsm}. The key idea of an iterative adversarial attack is to apply the FGSM perturbation $n$ times, changing the result of the previous iteration in every step $t$:
    \begin{align*}
        \vecX^{t+1}=\vecX^{t}+\epsilon \operatorname{sign}\left(\nabla_{\vecX^t} L\left(f(\vecX^t), y\right)\right).
    \end{align*}
    
    This method can help provide a more precise attack on the model by changing the direction of the attack for different data periods that shouldn't be transformed more for model compromising.

    \item \textbf{PGD attack}. The Projected Gradient Descent (PGD) \cite{madry2017towards} attack is a white-box universal “first-order adversary” method of attack. Unlike the iFGSM attack, the author proposes the idea of the limitation difference between adversarial data and original data objects. Perturbations generated in the $l_{\inf}$-ball $S$ of radius $\eta$ $T$ times with step size $\lambda$. In other words, if our perturbation reaches the border of the $l_{\inf}$-ball, we clip it. It helps to increase the robustness of the attack.  
    \begin{align*}
        \vecX^{t + 1}=\Pi_{\vecX + S}\left( \vecX^t +\lambda  \operatorname{sign}\left(\nabla_{\vecX^t} L\left(f(\vecX^t), y\right)\right)\right),
    \end{align*}
    where $\Pi_A$ is the projection operator for a set $A$.
    As in the original paper, we fix the step size $\lambda = \frac{2.5\eta}{T} $ to decrease the number of hyperparameters.
    
    \item \textbf{SimBA attack}. The SimBA is a simple but effective black box attack algorithm that modifies different parts of original objects. In the case of computer vision, it iteratively modifies different randomly chosen picture pixels. The algorithm was used for Cartesian coordinates. For computer vision objects, it means they take a randomly chosen colour of one pixel to transform it. Our variant of SimBA has two hyperparameters: the number of maximum iterations $T_{max}$ and the step size $\epsilon$. The algorithm of the attack, as well as its hyperparameters, are presented in Algorithm~\ref{alg:simba}. For a time series object, we will choose a one-time point and one feature in this time point if the data object is multivariate. 

\end{itemize}
    
    \begin{algorithm}
    \caption{SimBA attack}\label{alg:simba}
    \begin{algorithmic}
    \Procedure{SimBA}{$\vecX, y, Q, \epsilon, T_{max}$,}
    \State $\delta = 0$
    \State $p = p_f(y \ | \ \vecX)$
    \State $t = 0$
    \While{$p_y = \max_{y'} p_{y'}$ or $t < T_{max}$}
        \State Pick randomly without replacement: $q \in Q$
        \For{$\alpha \in \{ \epsilon,-\epsilon \} $}
            \State $p' = p_f(y \ | \ \vecX + \delta + \alpha q)$
            \If{$p'_y < p_y$}
                \State $\delta = \delta + \alpha q$
                \State $p' = p$
                
                \State break
            \EndIf
        \State $t = t + 1$
        \EndFor
    \EndWhile
    \State \textbf{return} $\delta$
    \EndProcedure
    \end{algorithmic}
    \end{algorithm}

\section{Considered aggregation functions}
\label{sec:app_aggregation}

We consider three reasonable aggregation functions: sum, harmonic, and hypercone.
The details for them are provided below.
For all methods, hyperparameter tuning was provided by a grid search of possible values for every hyperparameter with fixed attack number steps.

\begin{enumerate}
    \item \textbf{Sum aggregation}. This aggregation method was inspired by classical machine learning Ridge and LASSO regression loss when we take one loss as it is and add a second part with an extra $\alpha$ hyperparameter. An example of this loss for the iFGSM attack:
    $$
    g(\vecX) = a(\vecX) + \alpha d(\vecX) = L_{\mathrm{target}}\left(f(\vecX), y\right) - \alpha\log D (\vecX)
    $$
    \item \textbf{Harmonic aggregation}. Harmonic aggregation doesn't take some extra hyperparameters for tuning. The idea is to take the harmonic mean between the loss of the discriminator and the target model.
    \begin{align*}
    g(\vecX) &= \operatorname{mean_{harmonic}} (a(\vecX), d(\vecX)) = 2 \frac{a(x) \cdot d(x)}{a(x) + d(x)} = \\
    &= 2 \frac{L_{\mathrm{target}}\left(f(x), y \right) \cdot (- \log D (\vecX^t))}{L_{\mathrm{target}}\left(f(\vecX), y\right) - \log D (\vecX) + \gamma},     
    \end{align*}
    where $\gamma$ is a small value to avoid the zero division problems. 
    \item \textbf{Hypercone aggregation}. The idea behind this aggregation function is to project the discriminator loss gradient to the hypercone around the target attack gradient \cite{tercan2024thresholded}. The width of the hypercone depends on hyperparameter $\Delta$, which is the angel hypercone and axes. Increasing  $\Delta$ brings a greater role of regularization in the final vector. The main difference between this and other methods is that hypercone aggregation mixes final gradient vectors but not the loss values. So, the limitation of this method is the inability to apply hypercone regularization to the black box attacks.
    \begin{align*}
    \nabla_{\vecX}g(\vecX) &=  \frac{\cos\Delta}{\sin\phi} \sin(\Delta + \phi) 
    \Bigl( \nabla_{\vecX} L_{\mathrm{target}} + \\
    &+
    \nabla_{\vecX}L_{\mathrm{disc}} \frac{\|\nabla_{\vecX}L_{\mathrm{target}}\|}{\|\nabla_{\vecX}L_{\mathrm{disc}}\|} 
    \left( \sin\phi \tan\Delta - \cos\phi \right) \Bigr),
    \end{align*}
    where $\phi$ is the angle between the gradient of the target model loss $\nabla_{\vecX} L_{target}$ and the gradient of discriminator model loss $\nabla_{\vecX}L_{disc}$.
\end{enumerate}

\section{Comprehensive experiments results}
\label{secA1}

Table \ref{tab:all_metrics} presents the results of all main experiments. For most experiments, our regularised attacks outperform vanilla attacks and SGM attacks. 

\begin{table}[!thb]\centering
\caption{The Successfulness metric ($\uparrow$) of all experiments. The best attack for every dataset-model experiment is highlighted \textbf{bold} text style. Values marked with $^*$ are calculated with less strict rules for discriminator training. The reason is that it is impossible to train a discriminator with a quality above the specified threshold.}

\label{tab:all_metrics}
    \begin{tabular}{c|ccccc}
    \hline
    \multirow{2}{*}{Dataset} & \multirow{2}{*}{Attack} & 
    \multicolumn{4}{c}{Model} \\
    & & PatchTST & RNNA & ResCNN & S4 \\
    \hline
    \multirow{6}{*}{Coffee} & Vanilla iFGSM & 0.464 & 0.662 & \textbf{1.0} & 0.893 \\
     & Vanilla PGD & 0.133 & 0.339 & 0.036 & $0.018^*$  \\
     & SGM & 0.422 & \textbf{1.0} & 0.964 & 0.895 \\
     & PGD Sum Reg (our) & 0.557 & \textbf{1.0}  & \textbf{1.0}  & $0.861^*$  \\
     & iFGSM Harmonic Reg (our) & \textbf{1.0} & \textbf{1.0}  & \textbf{1.0}  & \textbf{1.0}  \\
     & iFGSM Sum Reg (our) & \textbf{1.0}  & \textbf{1.0} & \textbf{1.0} & \textbf{1.0} \\
    \hline
    \multirow{6}{*}{FreezerRegularTrain} & Vanilla iFGSM & 0.051 & 0.076 & 0.996 & 0.09 \\
     & Vanilla PGD & 0.537 & 0.464 & 0.648 & 0.742  \\
     & SGM & 0.457 & 0.243 & 0.992 & 0.46 \\
     & PGD Sum Reg (our) & \textbf{1.0} & 0.537 & 0.979 & 0.935  \\
     & iFGSM Harmonic Reg (our) & 0.995 & 0.469 & 0.994 & 0.85 \\
     & iFGSM Sum Reg (our) & \textbf{1.0} & \textbf{0.944} & \textbf{0.997} & \textbf{0.951} \\
    \hline
    \multirow{6}{*}{GunPoint} & Vanilla iFGSM & 0.02 & 0.013 & 0.275 & 0.294 \\
     & Vanilla PGD & $0.361^*$ & 0.179 & 0.304 & $0.517^*$ \\
     & SGM & 0.503 & $\mathbf{0.927^*}$ & 0.538 & 0.475 \\
     & PGD Sum Reg (our) & $\mathbf{0.814^*}$ & 0.469 & 0.575 & $0.704^*$ \\
     & iFGSM Harmonic Reg (our) & 0.767 & 0.46 & \textbf{0.859} & 0.864 \\
     & iFGSM Sum Reg (our) & 0.802 & 0.513 & 0.736 & \textbf{0.886} \\
    \hline
    \multirow{6}{*}{GunPointMaleVSFemale} & Vanilla iFGSM & 0.034 & 0.294 & 0.335 & 0.156 \\
     & Vanilla PGD & 0.132 & 0.382 & 0.162 & $0.323^*$ \\
     & SGM & 0.334 & $0.606^*$ & $0.928^*$ & $\mathbf{0.600^*}$ \\
     & PGD Sum Reg (our) & 0.541 & \textbf{0.705} & 0.579 & $0.441^*$ \\
     & iFGSM Harmonic Reg (our) & 0.501 & 0.523 & 0.778 & 0.36 \\
     & iFGSM Sum Reg (our) & \textbf{0.673} & 0.415 & \textbf{1.0} & \textbf{0.581} \\
    \hline
    \multirow{6}{*}{PowerCons} & Vanilla iFGSM & 0.187 & 0.376 & 0.328 & 0.404 \\
     & Vanilla PGD & 0.187 & 0.265 & 0.575 & 0.277 \\
     & SGM & 0.199 & 0.118 & 0.401 & $0.412^*$ \\
     & PGD Sum Reg (our) & 0.543 & \textbf{0.522} & 0.66 & 0.281 \\
     & iFGSM Harmonic Reg (our) & 0.869 & 0.46 & \textbf{0.931} & 0.689 \\
     & iFGSM Sum Reg (our) & \textbf{0.97} & 0.379 & 0.636 & \textbf{0.815} \\
    \hline
    \multirow{6}{*}{Strawberry} & Vanilla iFGSM & 0.053 & 0.0 & 0.078 & 0.12 \\
     & Vanilla PGD & 0.027 & 0.003 & 0.037 & 0.172 \\
     & SGM & 0.654 & 0.576 & 0.544 & 0.427 \\
     & PGD Sum Reg (our) & 0.665 & 0.577 & 0.651 & 0.522 \\
     & iFGSM Harmonic Reg (our) & \textbf{0.921} & \textbf{0.943} & \textbf{0.965} & 0.745 \\
     & iFGSM Sum Reg (our) & 0.769 & 0.537 & 0.45 & \textbf{0.799} \\
    \hline
    \end{tabular}
\end{table}



\section{Examples of attacked data samples}
\label{sec:data_samples}

\begin{figure}[ht]
\centering
\begin{subfigure}{.49\textwidth}
  \centering
  \includegraphics[width=.97\linewidth]{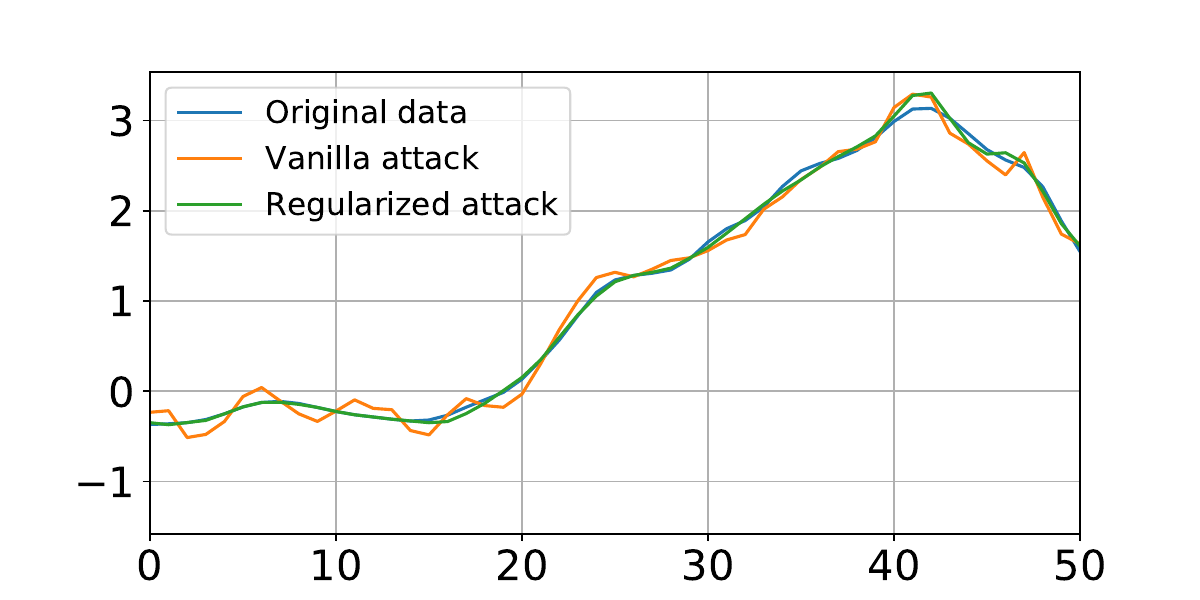}
  \caption{Strawberry, ResCNN}
  \label{fig:sample1}
\end{subfigure}
\begin{subfigure}{.49\textwidth}
  \centering
  \includegraphics[width=.97\linewidth]{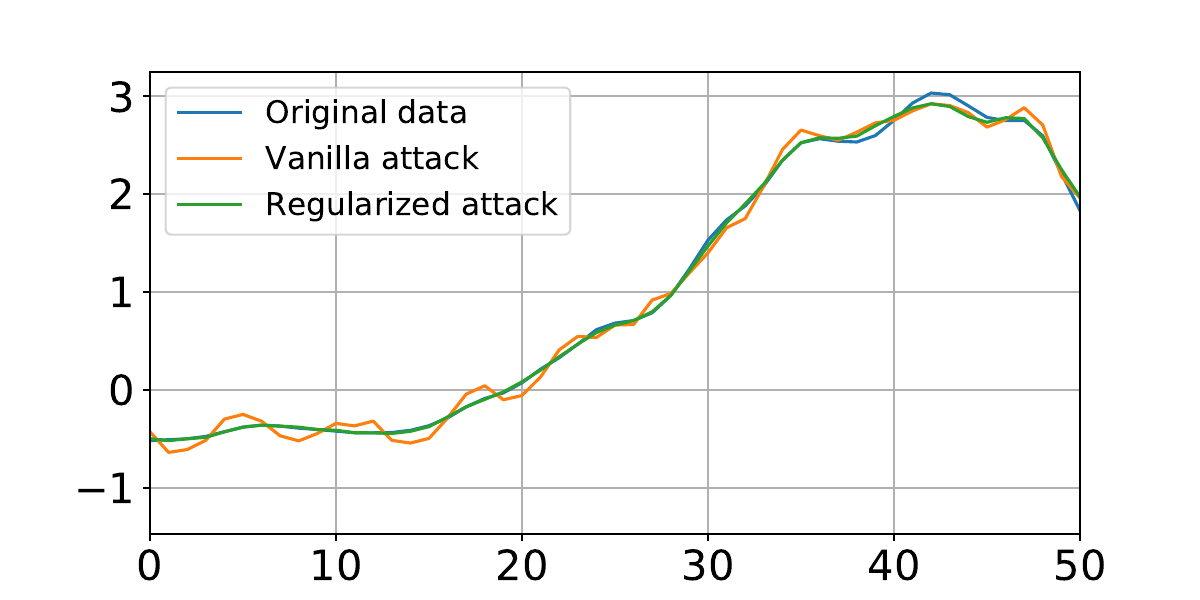}
  \caption{Strawberry, ResCNN}
  \label{fig:sample2}
\end{subfigure}
\begin{subfigure}{.49\textwidth}
  \centering
  \includegraphics[width=.97\linewidth]{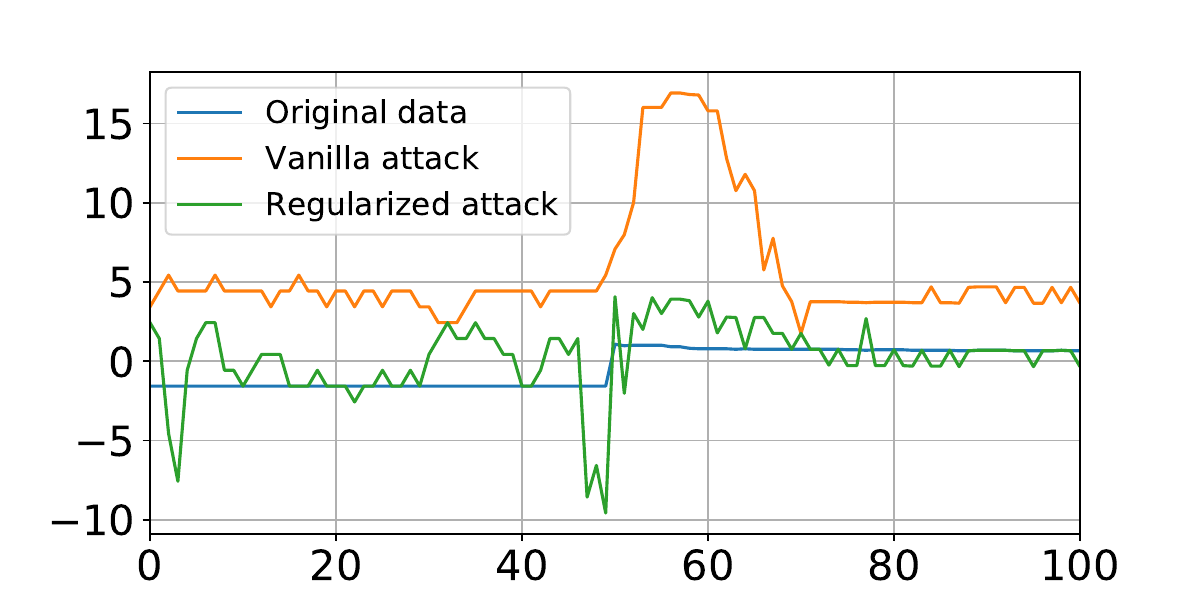}
  \caption{FreezerRegularTrain, RNNAttention}
  \label{fig:sample3}
\end{subfigure}
\begin{subfigure}{.49\textwidth}
  \centering
  \includegraphics[width=.97\linewidth]{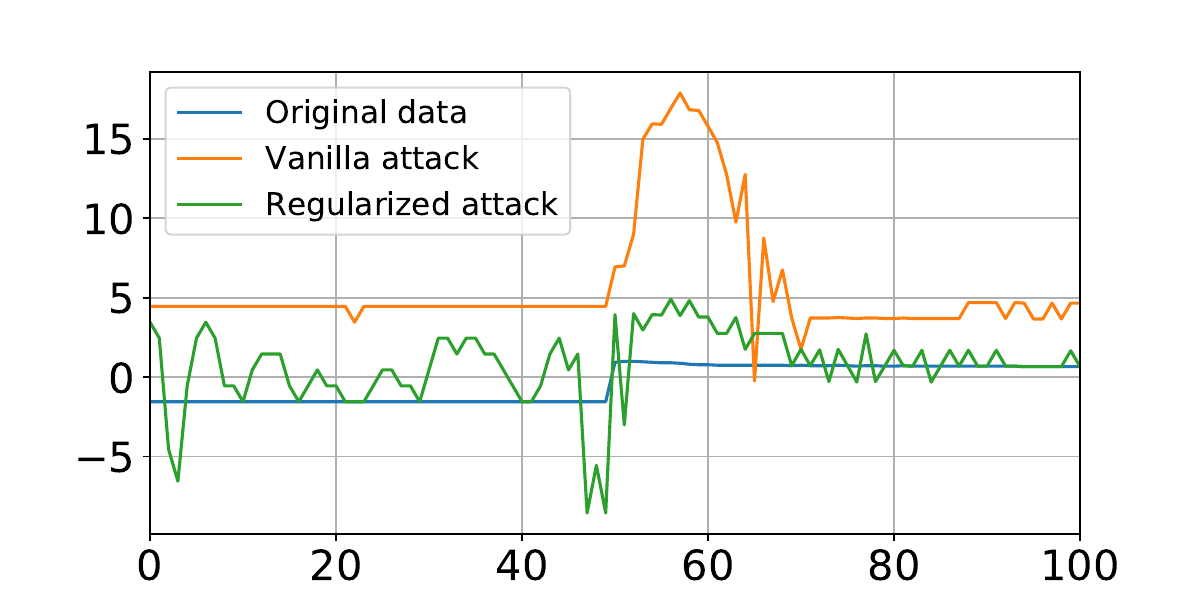}
  \caption{FreezerRegularTrain, RNNAttention}
  \label{fig:sample4}
\end{subfigure}
\begin{subfigure}{.49\textwidth}
  \centering
  \includegraphics[width=.97\linewidth]{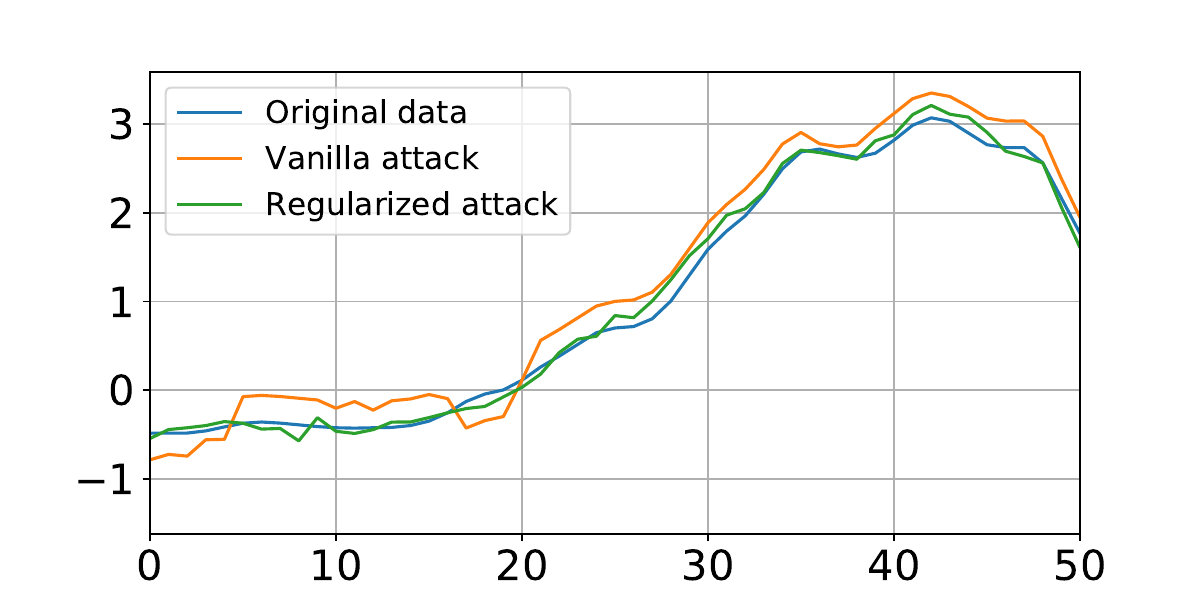}
  \caption{Strawberry, PatchTST}
  \label{fig:sample5}
\end{subfigure}
\begin{subfigure}{.49\textwidth}
  \centering
  \includegraphics[width=.97\linewidth]{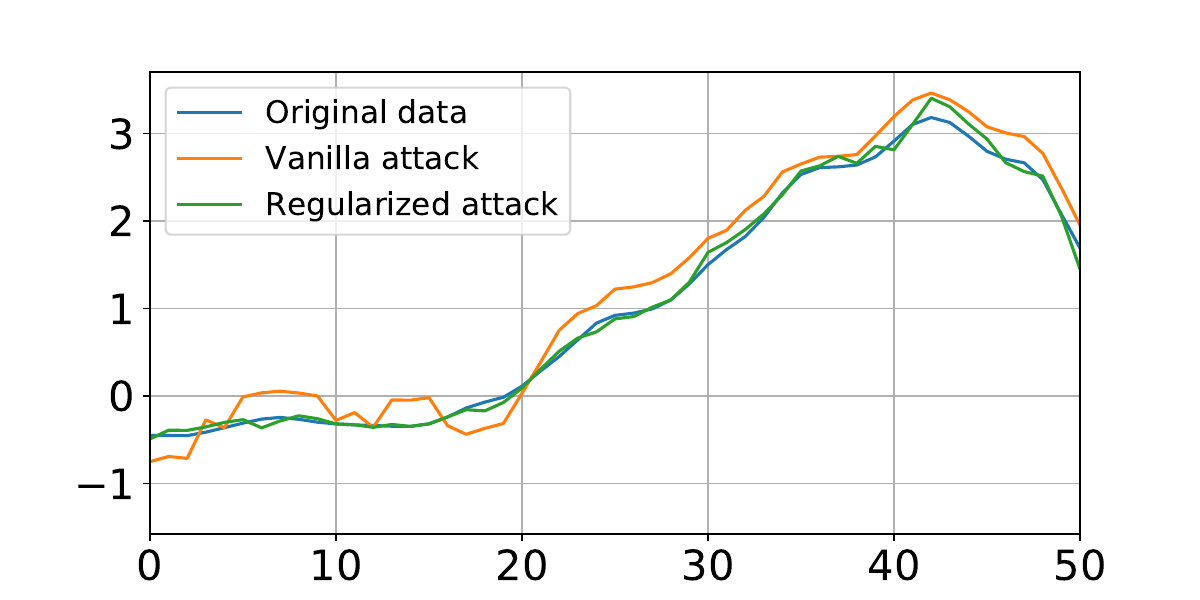}
  \caption{Strawberry, PatchTST}
  \label{fig:sample6}
\end{subfigure}
\begin{subfigure}{.49\textwidth}
  \centering
  \includegraphics[width=.97\linewidth]{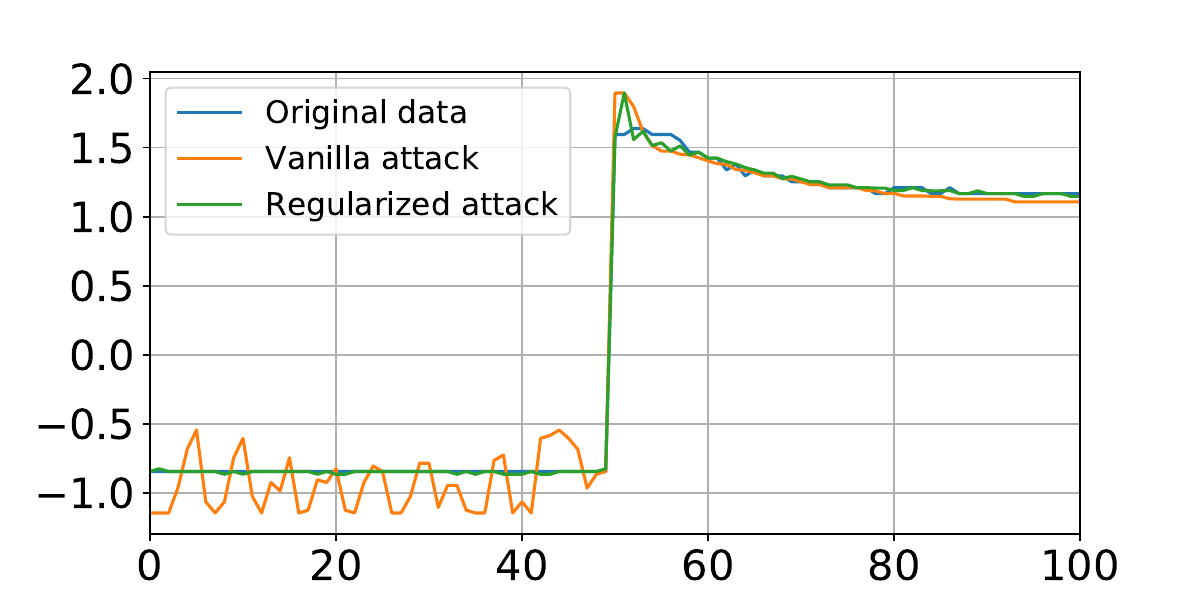}
  \caption{FreezerRegularTrain, ResCNN}
  \label{fig:sample7}
\end{subfigure}
\begin{subfigure}{.49\textwidth}
  \centering
  \includegraphics[width=.97\linewidth]{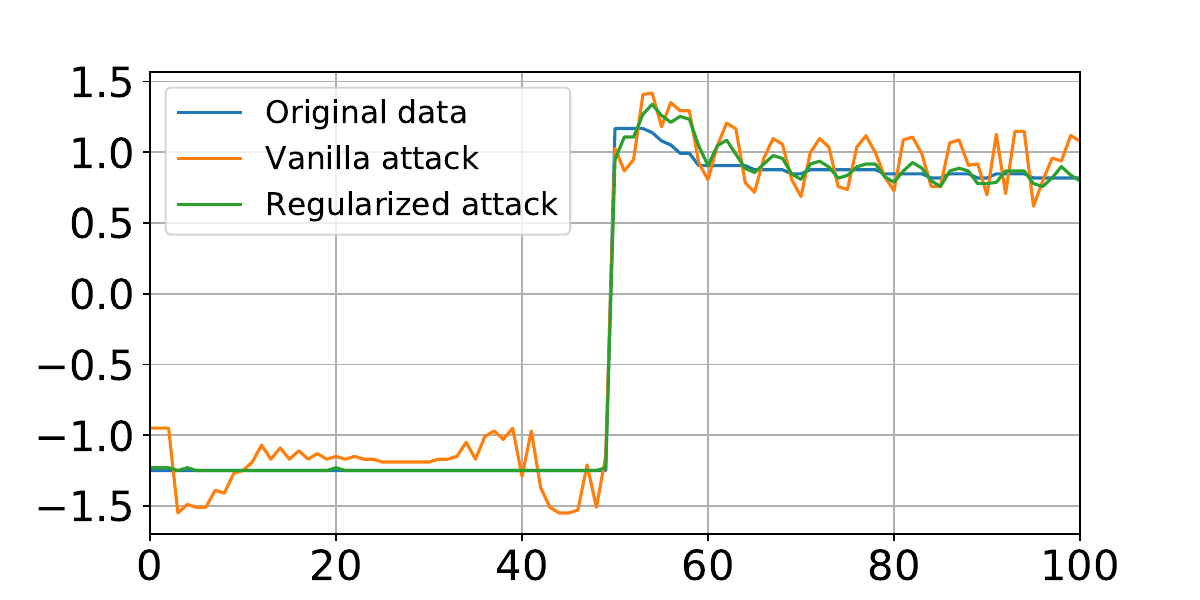}
  \caption{FreezerRegularTrain, ResCNN}
  \label{fig:sample8}
\end{subfigure}
\caption{Examples of iFGSM attacks on various models and datasets, both with and without regularization.}
\label{fig:samples}
\end{figure}

Figure~\ref{fig:samples} presents examples of both the vanilla iFGSM attack and the iFGSM attack with sum regularization. These examples illustrate scenarios in which both attack variants successfully altered the target model's predictions. However, the regularized attack also remained undetected, as demonstrated by misclassification from the discriminator model. Both attack variants were applied with equal strength, while the regularization parameter $\alpha$ was tuned to ensure concealment.

For better visual clarity, the time-series data has been truncated to display only the first 50 or 100 points, depending on the dataset.

While both attacks resulted in a change to the model’s class predictions, the discriminator scores for the vanilla attack were close to 1.0, indicating detection. In contrast, the regularized attack consistently produced discriminator scores below 0.2, indicating that the discriminator failed to detect any perturbations. The regularized attack produces fewer artifacts and generates samples that more closely resemble the original data, as evidenced by its ability to deceive the discriminator model.

\end{appendices}



\end{document}